\documentclass{article}

\usepackage{spconf,amsmath,epsfig}
\usepackage{amssymb}
\usepackage{amsthm}
\usepackage{algorithm}
\usepackage{algpseudocode}
\usepackage{tikz}
\usepackage{tabu}
\usepackage{graphicx}
\usepackage{subfig}
\usepackage{epstopdf}
\usepackage{epsfig}
\usetikzlibrary{tikzmark,calc}
\usepackage{float}
\usepackage{multirow}
\usepackage{hyperref}
\numberwithin{equation}{section} 

\usepackage{microtype}

\newcommand\Algphase[1]{%
\vspace*{-.7\baselineskip}\Statex\hspace*{\dimexpr-\algorithmicindent-2pt\relax}\rule{\columnwidth}{0.4pt}%
\Statex\hspace*{-\algorithmicindent}\textbf{#1}%
\vspace*{-.7\baselineskip}\Statex\hspace*{\dimexpr-\algorithmicindent-2pt\relax}\rule{\columnwidth}{0.4pt}%
}

\DeclareMathOperator*{\argmin}{arg\,min}

\title{Reducing the Search Space for Hyperparameter \\ Optimization Using Group Sparsity}
\name{Minsu Cho and Chinmay Hegde\thanks{Email: \{chomd90,chinmay\}@iastate.edu. This work is supported in part by grants from NSF CCF-1750920, a Faculty Fellowship from the Black and Veatch Foundation, and an equipment donation from NVIDIA Corporation.}
\address{Iowa State University \\ ECE Department \\ Ames, IA, USA 50011}}
\begin{document}
\ninept
\maketitle
\begin{abstract}

We propose a new algorithm for hyperparameter selection in machine learning algorithms. The algorithm is a novel modification of Harmonica, a spectral hyperparameter selection approach using sparse recovery methods. In particular, we show that a special encoding of hyperparameter space enables a natural group-sparse recovery formulation, which when coupled with HyperBand (a multi-armed bandit strategy) leads to improvement over existing hyperparameter optimization methods such as Successive Halving and Random Search. Experimental results on image datasets such as CIFAR-10 confirm the benefits of our approach.
\end{abstract}

\section{Introduction}
\label{sec:intro}

\subsection{Setup}

Machine learning (ML) models have been developed successfully to perform complex prediction tasks in recent years. However, most ML algorithms, especially in deep learning, require manual selection of several hyperparameters such as learning rate, regularization penalty constants, dropout ratio, and model architecture. The quality of the model depends on how the designer of the ML model has carefully chosen the hyperparameters; however, the complexity and variety of ML models magnifies the practical difficulties of selecting appropriate combinations of parameters to maximize performance. The area of hyperparameter optimization (HPO) addresses the problem of searching for the optimal choices in hyperparameter space.

Formally, let $X$ denote the space of hyperparameters (whether numerical and categorical), and let $f$ be the function mapping from $X$ to the test loss obtained by training a given ML algorithm with a particular set of hyperparameters. The goal of HPO is to approximate a set of hyperparameters ``close enough'' to the global optimum 
$$x^* = \argmin_{x \in X}f(x)$$
as efficiently as possible. 

\subsection{Prior Work}
Traditionally, ML practitioners have solved the HPO problem via brute-force techniques such as grid search over $X$. This strategy quickly runs into exponentially increasing computation costs with each additional dimension in hyperparameter space. As a solution, Bayesian Optimization (BO) techniques have been proposed. These assume a certain prior distribution over the cost function $f(x)$ and updates the posterior distribution with each new ``observation'' (or measurement of training loss) at a given set of hyperparameters~\cite{bergstra2011algorithms, hutter2011sequential, snoek2012practical, thornton2013auto, eggensperger2013towards, snoek2014input, ilievski2017efficient}. Subsequently, an acquisition function samples the posterior to form a new set of hyperparameters, and the process iterates.

Despite the popularity of BO techniques, they often provide unstable performance, particularly in high-dimensional hyperparameter space. An alternative technique to BO is Random Search (RS), which not only provides computational efficiency compared to grid search, but also strong ``anytime'' performance with easy parallel implementation~\cite{bergstra2012random}. 

Multi-armed bandit (MAB) approaches adapt the random search strategy to allocate the different resources to the randomly chosen candidate points to speed up the convergence to the optimum instead of spending full resources as random search and the BO. Successive Halving (SH) and Hyperband adapt the multi-armed bandit approach to random search, picking more candidates than random search with the same amount of budget by pruning poorly-performing hyperparameters in the early state~\cite{jamieson2016non,li2017hyperband,kumar2018parallel}. 
In contrast with BO techniques (which are hard to parallelize), the integration of BO and Hyperband achieve both advantages of guided selection and parallelization~\cite{wang2018combination, falkner2018bohb, bertrandhyperparameter}.

Gradient descent methods~\cite{bengio2000gradient,maclaurin2015gradient,luketina2015scalable,fu2016drmad, franceschi2017forward} (or more broadly, meta-learning approaches) have also been applied to solve the HPO problem, but these are only suitable to optimize continuous hyperparameters. Since this is a very vast area of current research, we do not compare our approach with these techniques.

While BO dominates the model-based approach, a recent technique called \emph{Harmonica} proposed a \emph{spectral} approach, applying ideas from \emph{sparse recovery} on a Boolean version of the objective function. Using this approach, Harmonica provides the unique benefit of reducing the dimensionality of hyperparameter space by quickly finding highly influential hyperparameters, following which other standard (search or optimization) techniques can be used~\cite{hazan2017hyperparameter}.

\subsection{Our Contributions}

Our main contribution is an extension to the Harmonica algorithm. While it successfully demonstrates finding important categorical features, we focus on finding the numerical features by proposing a new representation on numerical hyperparameter values. The representation not only reduces the dimension of hyperparameter space, but also groups the hyperparameters based on knowledge of its structure to achieve improved accuracy and stability. 

To supplement our algorithm, we validate our numerical expression with hyperparameters grouping to examine its guidance accurately. We visually show that this algorithm closely approximates the global minimum in hyperparameter space by plotting the loss surface with two hyperparameters. We also show the robustness of our proposed algorithm combining the guidance to the decision-theoretic methods with measurable improvements in test loss using a CNN architecture trained on the CIFAR-10 image classification dataset. 

\subsection{Techniques}

Following~\cite{falkner2018bohb}, we observe three desiderata to be satisfied with any HPO method: parallelizability, scalability, and strong final performance. The first criterion is parallelizability of the algorithm since HPO requires expensive computations. We use Hyperband, which is the current state-of-art in multi-armed bandit approaches, as the base algorithm to satisfy the first qualification. 

To achieve the second and third criteria, we use the Harmonica trick~\cite{hazan2017hyperparameter}: we first binarize the hyperparameter space, and decompose the Fourier expansion of the (Boolean) function $f$. Finding the influential hyperparameters from a small number of (sampled) training loss observations reduces to solving a group-sparse recovery problem from compressive measurements. This leads us to better overall test error for a given computational budget.

\section{Mathematical model and Algorithm}
\label{sec:alg}

We now present our HPO algorithm; we restrict our attention to discrete domains (and assume that continuous hyperparameters have been appropriately binned). Let $f: \{-1, 1\}^n \mapsto \mathbb{R}$ be the loss function to be optimized. Let there be $k$ different types of hyperparameters. In other words, we allocate $n_i$ bits to the $i^{\textrm{th}}$ hyperparameter category such that $\sum_{i=1}^{k} n_i=n$. The task of HPO involves searching the approximate hyperparameters close to the global minimizer 
\begin{align}
    x^* = \argmin_{x\in\{-1,1\}^n}f(x) .
\end{align}

\subsection{PGSR-HB}\label{PGSR-HB}

We propose \emph{Polynomial Group-Sparse Recovery within Hyperband} (PGSR-HB), a new HPO search algorithm which enables considerable reduction of the hyperparameter space. We combine Hyperband, the multi-armed bandit method that balances exploration and exploitation from uniformly random sampled hyperparameter configurations, with a group sparse version of Polynomial Sparse Recovery, which is the main component of the spectral decomposition-based Harmonica method of HPO. Algorithm~\ref{alg:PGSRHB} shows the pseudo code of PGSR-HB.

\begin{algorithm}[t]
\caption{\textsc{PGSR-HB}}
\label{alg:PGSRHB}
\begin{algorithmic}[1]
\State\textbf{Inputs:} Resource $R$, scaling factor $\eta$, total cycle $c$ 
\State\textbf{Initialization:} $s_{max}=\lfloor \log_{\eta}(R)\rfloor$, $B=(s_{max}+1)R$, input history $H_{input}=\emptyset$, output history $H_{output}=\emptyset$
\For{$round=1:c$}
\For{$s \in \{s_{max}, s_{max}-1,\ldots,0\}$}
\State $n=\lceil \frac{B}{R} \frac{\eta^s}{(s+1)}\rceil$, $r=R\eta^{-s}$
\State T = {\fontfamily{pcr}\selectfont PGSR\_Sampling(n)}
\For{$i \in \{0,\ldots,s\}$}
\State $n_i = \lfloor n\eta^{-i}\rfloor$
\State $r_i = r\eta^i$
\State $L = \{f(t,r_i): t \in T\}$
\State $H_{input,r_i} \leftarrow H_{input,r_i} \bigcup T$
\State $H_{output,r_i} \leftarrow H_{output,r_i} \bigcup L$
\State $T = sort(T,L,\lfloor \frac{n_i}{\eta} \rfloor)$
\EndFor
\EndFor
\EndFor
\State\textbf{return} Configuration with the smallest loss
\Algphase{Sub-algorithm - PGSR Sampling}
\State\textbf{Input:} $H_{input}$, $H_{output}$, sparsity $s$, polynomial degree $d$, minimum observations $T$, randomness ratio $\rho$
\If{every $|H_{output,r}| < T$} \Return random sample from original domain of $f$.
\EndIf
\State Pick $H_{input,r}$ and $H_{output,r}$ with largest $r$: $|H_{output,r}|\geq T$.
\State Group Fourier basis based on hyperparameter structure.
\State Solve $$x^* = \argmin_{\alpha} \frac{1}{2}\|y - \sum_{l=1}^{m}\Psi^{l} \alpha^{l}\|_{2}^{2} + \lambda \sum_{l=1}^{m} \sqrt{p_{l}} \|\alpha^{l}\|_2$$
\State Let $S_1, \ldots S_s$ be the indices of the largest coefficient of $\alpha$. Then, $g(x) = \sum_{i \in [s]} \alpha_{S_i} \chi_{S_i}(x)$ and $J = \bigcup_{i=1}^{s}S_i$
\State With probability $\rho$, return random sample from original domain of $f$; else return random sample from reduced domain of $f_{J,x^{*}}$.
\end{algorithmic}
\end{algorithm}

PGSR-HB adopts the decision-theoretic approach of Hyperband, but with the additional features of tracking the history of all loss values from different resources. Hyperband contains the subroutine algorithm, Successive Halving (abbreviated as SH, see Lines 7-14), following the assumption that the performance of different hyperparameter choices in the process of training indicates which configurations are worth investing further resources, and which ones are fit to discard.

Let $R$ denote the (units of computational) resource to be invested in one round to observe the final performance of the model; $\eta$ denote a scaling factor; and $c$ the total number of rounds. Defining $s_{max}=\log_{\eta}R$, the total budget spent from SH is $B=(s_{max}+1)R$. The algorithm samples $n$ configurations with a sub-routine (which we call \emph{PGSR-Sampling}, and explain further in the next section). Here, $n$ is given by:
\begin{align}
    n = \lceil \frac{B}{R} \frac{\eta^s}{(s+1)}\rceil
\end{align}
and calculate the test loss with 
\begin{align}
    r = R\eta^{-s}
\end{align}
epochs of training. The function $f(t,r_i)$ in Algorithm~\ref{alg:PGSRHB} (Line 10) returns the intermediate test loss of a hyperparameter configuration $t$ with $r_i$ of training epochs. Since the test loss is the metric to measure the performance of the model, the algorithm keeps only the top $\frac{1}{\eta}$ configurations (Line 13) and repeats the process by increasing the training epochs by the factor of $\eta$ until $r$ reaches to resource $R$. While SH introduces the new hyperparameter $s$, SH aggressively explores the hyperparameter space as $s$ close to $s_{max}$ while SH with $s$ equal to zero is equivalent to random search (aggressive exploitation). The algorithm with one cycle contains $(s_{max}+1)$ subroutines of SH attempting different levels of exploration and exploitation with all possible $s$ values (Line 4).

\subsection{PGSR Sampling}

As PGSR-HB collects the outputs of the function $f$, the PGSR-Sampling sub-routine recovers Fourier basis coefficients of the Boolean function $f$ using techniques from sparse recovery to reduce the hyperparameter space. Before we discuss about how PGSR Sampling works and compare differences with Polynomial Sparse Recovery in the Harmonica method of~\cite{hazan2017hyperparameter}, we first establish some standard concepts in Fourier analysis of Boolean functions~\cite{o2014analysis}. Consider a function $f$ defined from $\{-1, 1\}^n$ to $\mathbb{R}$. The Fourier basis corresponding to any subset of indices $S$ (such that $S \subseteq [n]$) is defined as
\begin{align}
\label{eq:Fourier Basis}
\chi_S(x) = \prod_{i \in S} x_i
\end{align}
where $x_i$ is the $i^{th}$ element of the input vector. Then, the function $f$ can uniquely expressed as the series of a real multilinear polynomial basis (or Fourier basis) given by:
\begin{align}
\label{eq:Fourier Series}
f(x) = \sum_{S \subseteq [n]} \hat{f}(S) \chi_S(x)
\end{align}
where
\begin{align}
\hat{f}(S) = \mathbb{E}_{x\in\{-1,1\}^n}[f(x)\chi_S(x)]
\end{align}
where the expectation is taken with respect to the uniform distribution over the nodes of the $n$-dimensional hypercube. The \emph{restriction}~\cite{o2014analysis} of the Boolean function $f$ by a restriction pair $(J,z)$ where $J \subseteq [n]$ and $z\in\{-1,1\}^J$ is denoted by the function $f_{J,z}$ over $n-|J|$  variables by fixing the variables in $J$ to $z$. 

While Harmonica does not explicitly address how to discretize continuous hyperparameters, we introduce a simple mathematical expression that efficiently induces additional sparsity in the Fourier representation of $f$. Let $x$ be the $m$-digit binary number mapping to the set of integers with cardinality $2^m$ by function $g$, and $y$ be the $n$-digits binary number mapping to the set of numbers with cardinality $2^n$ which are evenly spaced in (0,1] by function $h$. Then we express the $i^{th}$ numerical hyperparameter value $hp_i$, for all $k$ categories ($i=1,\ldots,k$), in a log-linear manner as follows:
\begin{align}
\label{eq:HpRepresentation}
    hp_i = 10^{g(x)} \cdot h(y)
\end{align}

Our experimental results section shows how this simple nonlinear binning representation induces sparsity on function $g$, which captures the value's order of magnitude. As PGSR returns the features regard to the function $g$, the new representation efficiently reduces the hyperparameter space. While PSR in Harmonica recovers the Boolean function with Lasso~\cite{tibshirani1996regression}, the intuitive extension (arising from the above log-linear representation) is to replace sparse recovery with Group Lasso~\cite{yuan2006model}; this is used in Algorithm~\ref{alg:PGSRHB} (Line 23) as we group them based on the $g$ and $h$ based on hyperparameter categories. Let $y\in \mathbb{R}^m$ be the observation vector; let the hyperparameters be divided into $m+n$ groups (corresponding to functions $g$ and $h$) and let $\Psi^{l}$ is the submatrix of $\Psi \in \mathbb{R}^{m \times \binom{m}{d}}$ where its columns match the $l^{th}$ group. Similarly, $\alpha^{l}$ is a weight vector corresponding to the submatrix $\Psi^{l}$ and $p_{l}$ be the length of vector $\alpha^{l}$. In order to construct the submatrices which are the collection of Fourier basis on its columns by the hyperparameter structure, let there exist a set of groups $G = \{g_1, \ldots, g_m, h_1, \ldots, h_n\}$ as defined above. If there are $k$ possible combinations of groups from $G$ such that a $d$-degree Fourier basis exists, we derive the $k$ submatrices $\Psi^{1}, \ldots, \Psi^{k}$ using Eq. ~\eqref{eq:Fourier Basis}. Then the problem becomes equivalent to a convex optimization problem known as the Group Lasso, represented by the equation:
\begin{align}
\label{eq:GroupLasso}
    \min_{\alpha} \frac{1}{2}\|y - \sum_{l=1}^{m}\Psi^{l} \alpha^{l}\|_{2}^{2} + \lambda \sum_{l=1}^{m} \sqrt{p_{l}} \|\alpha^{l}\|_2
\end{align}

Lastly, the algorithm requires the input $\rho$ which represents a reset probability parameter that produces random samples from the original reduced hyperparameter space. This parameter prevents gathering the biased observations in different PGSR stages, since the measurements with substantial resources mostly arise from the later stages of Successive Halving. 

\subsection{Differences between PGSR-HB and Harmonica}
The standard Harmonica method samples the measurements under a uniform distribution before starting the search algorithm to recover the function $f$ with PSR (the sparse recovery through $l_1$ penalty, or standard Lasso). Harmonica requires ML designers to choose the number of randomly sampled measurements and its resources (training epochs) before starting the search algorithm. The reliability of measurements, especially in the deep learning literature, hugely depends on the number of resources used on each sampled point. Investing enormous resources in recovering Fourier coefficients guarantees that the Lasso regression performs reliably, but this is inefficient with respect to total budget; however, collecting the measurements with small resources would make PSR fail to provide the correct guidance for the outer search algorithm. We have experimented with other penalties than the standard L1-penalty: for example, Tikhonov regularization prevents model overfitting particularly in deep architectures. However, the regularized regression tends to learn slower than the model without a regularization, consequently misleading the search algorithm with the worst performance. Since PGSR-HB gathers all the function outputs -- from cheap resources to the most expensive resources -- PGSR-HB eliminates the need to set an explicit number of samples and training epochs as in Harmonica. 

The experimental results in \cite{hazan2017hyperparameter} shows significant promise in finding the influential categorical hyperparameters such as presence/absence of the Batch-normalization layer, or determining the descent algorithm (stochastic gradient descent vs. Adam) [both of which can be represented using binary variables], but limitations in optimizing the numerical hyperparameters such as learning rate, weight decay $l_2$ penalty, and batch size. PGSR-HB overcomes this limitation of Harmonica with the log-linear representation capturing both order-of-magnitude and details in~\eqref{eq:HpRepresentation} and Group Lasso~\eqref{eq:GroupLasso}.

\section{Experimental Results}
\label{sec:result}

\begin{figure}[b]
\begin{center}
    \setlength{\tabcolsep}{.1pt}
    \renewcommand{\arraystretch}{.1}
    \begin{tabular}{cc}
    \includegraphics[trim = 30mm 105mm 77mm 105mm, clip, width=0.8\linewidth]{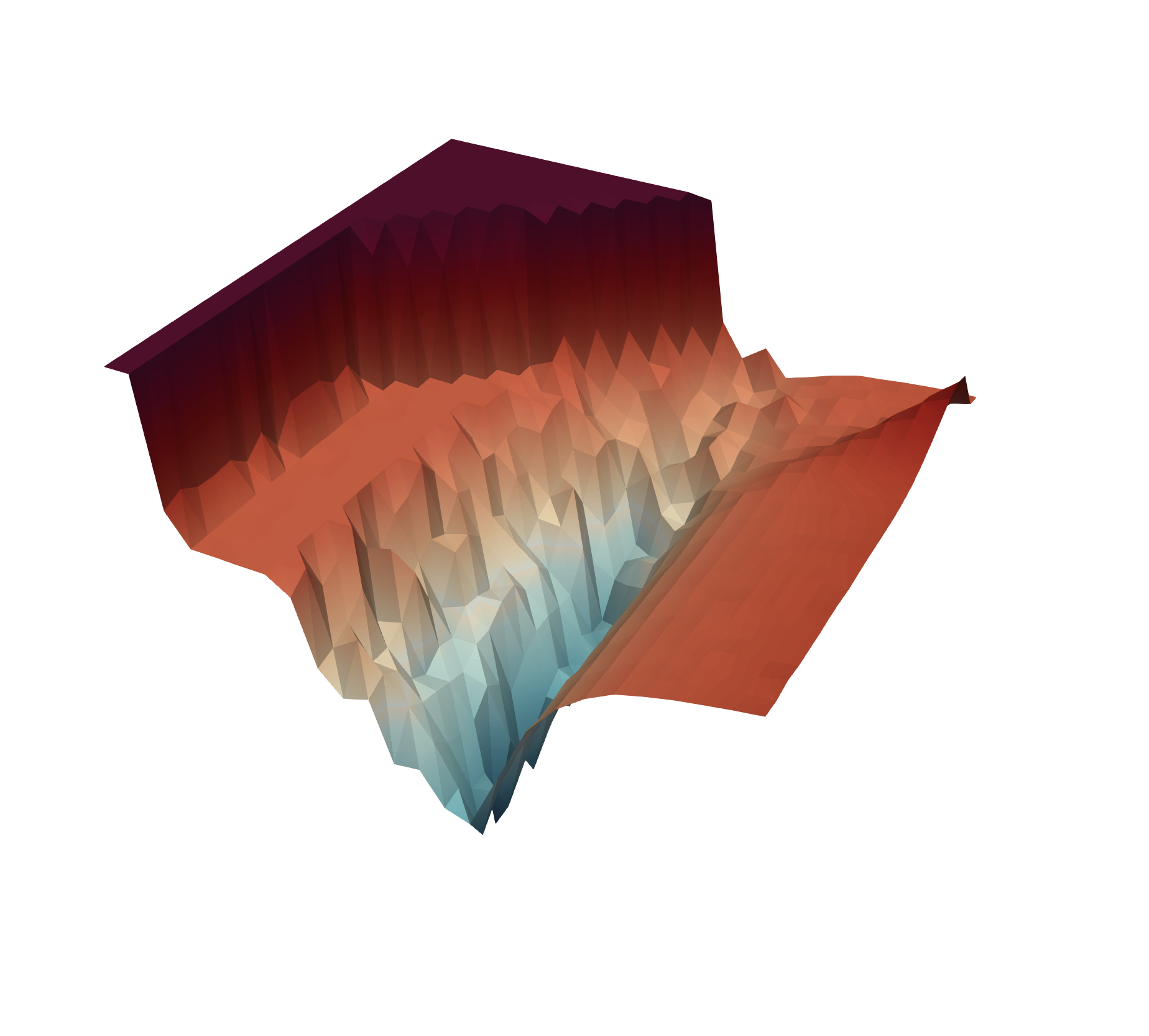}
    \end{tabular}
\end{center}
\caption{\emph{Test loss surface with two hyperparameters. Learning rate vs conv1 l2 penalty.}}
\label{fig:lossSurface}
\end{figure}

We verify the robustness of PGSR-HB by generating a test loss surface picking two hyperparameter categories as shown in Figure~\ref{fig:lossSurface}. We calculate the test loss by training 120 epochs with the standard benchmark image classification dataset, CIFAR-10. We used the convolutional neural network architecture from the cuda-convnet-$82\%$ model that has been used in previous work (\cite{jamieson2016non} and \cite{li2017hyperband}). We specifically choose the range of learning rate and the weight-decay penalty on the first convolutional layer to be from $10^{-6}$ to $10^{2}$. We keep the log scale with base ten on both horizontal and vertical axis to visualize the loss surface with more natural interpretation and dynamic range on the test loss. 

\begin{figure}[t]
\begin{center}
    \setlength{\tabcolsep}{.1pt}
    \renewcommand{\arraystretch}{.1}
    \begin{tabular}{cc}
    \includegraphics[trim = 35mm 40mm 35mm 40mm, clip, width=.8\linewidth]{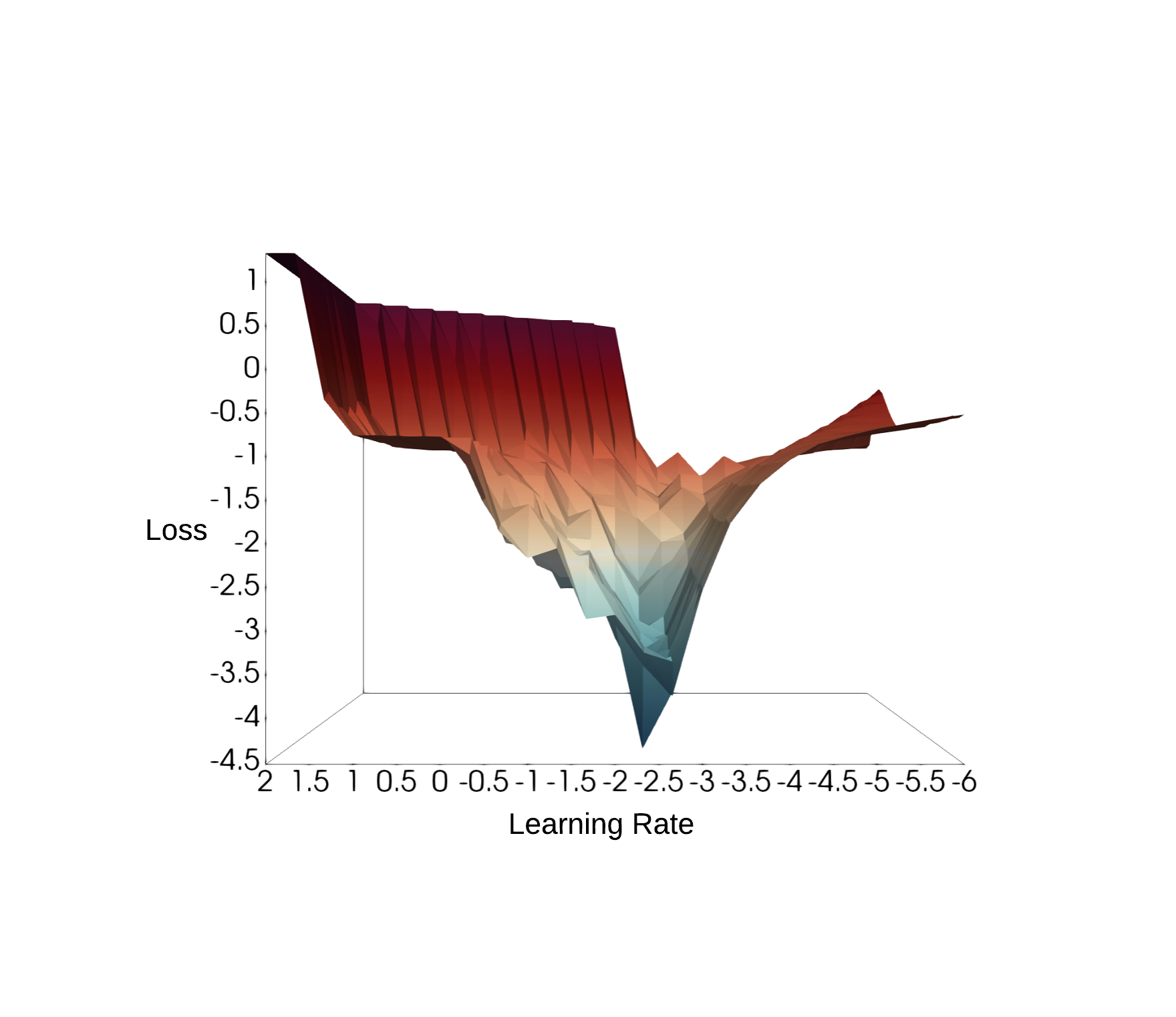}
    \end{tabular}
\end{center}
\caption{\emph{The view from learning rate axis.}}
\label{fig:lossSurfaceXaxis}
\end{figure}

\begin{figure}[t]
\begin{center}
    \setlength{\tabcolsep}{.1pt}
    \renewcommand{\arraystretch}{.1}
    \begin{tabular}{cc}
    \includegraphics[trim = 35mm 40mm 35mm 40mm, clip, width=0.8\linewidth]{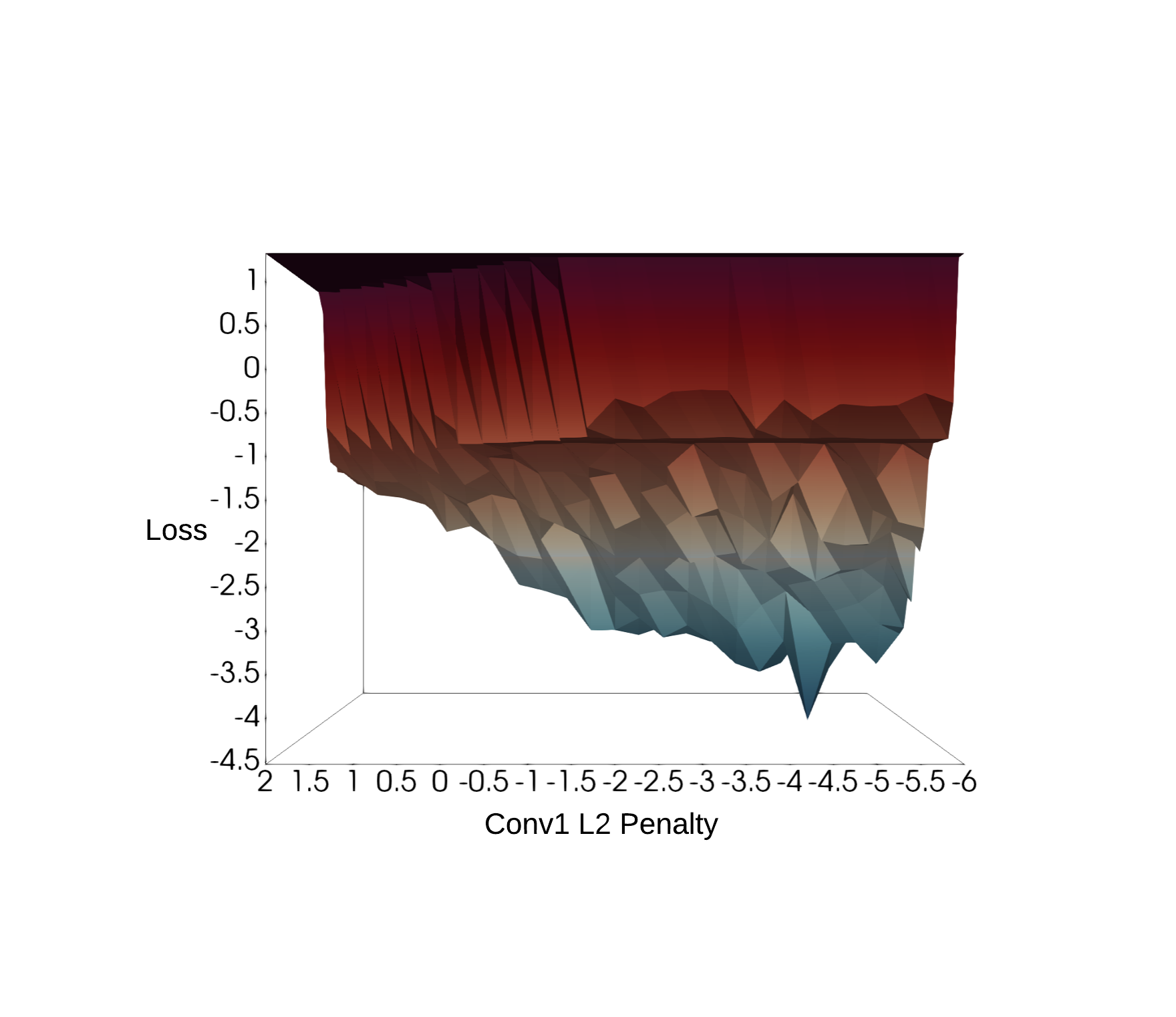}
    \end{tabular}
\end{center}
\caption{\emph{The view from conv1 l2 penalty axis.}}
\vspace{0.5cm}
\label{fig:lossSurfaceYaxis}
\end{figure}

Table~\ref{table:LrvsConv1} compares the performance of PGSR and PSR with ~\eqref{eq:HpRepresentation}, and PSR with evenly spaced hyperparameter values in log scale. The third and fourth columns in Table~\ref{table:LrvsConv1} list the reduced hyperparameter space for learning rate and first convolution layer l2 penalty by each algorithm. The experiment result shows that~\eqref{eq:HpRepresentation} induces improved sparsity to reduce the space further than the conventional method. Giving extra information of the hyperparameter structure with grouping not only helped PGSR to return the correct guidance, but also provided the stability on the lasso coefficient $\lambda$ as shown in the test loss surfaces (Figure~\ref{fig:lossSurfaceXaxis} and Figure~\ref{fig:lossSurfaceYaxis}) with PGSR results in Table~\ref{table:LrvsConv1}. More results of PGSR guidance with loss surfaces are in \href{https://chomd90.github.io/}{https://chomd90.github.io/}.

\begin{table}[!t]
\caption{Guidance Comparison on Learning Rate and Conv1 L2}
\label{table:LrvsConv1}
\begin{center}
\begin{tabular}{|c|c|c|c|c|c|}
\hline
Method &$\lambda$& Learn Rate & Conv1 Penalty\\
\hline
PGSR & $0.5$ & $[\mathbf{10^{-3}, 10^{-2}}]$ & $[\mathbf{10^{-5}, 10^{-4}}]$ \\
\hline
PGSR & $1.0$ & $[\mathbf{10^{-3}, 10^{-2}}]$ & $[\mathbf{10^{-5}, 10^{-4}}]$ \\
\hline
PGSR & $2.0$ & $[\mathbf{10^{-3}, 10^{-2}}]$ & $[\mathbf{10^{-5}, 10^{-4}}]$ \\
\hline
PSR & $0.5$ & $[\mathbf{10^{-3}, 10^{-2}}]$ & $[10^{-6}, 10^{2}]$ \\
\hline
PSR & $1.0$ & $[10^{-4}, 10^{-3}]$ & $[10^{-3}, 10^{-2}]$ \\
\hline 
PSR & $2.0$ & $[10^{0}, 10^{2}]$ & $[10^{-3}, 10^{-2}]$ \\
\hline
PSR w/o~\eqref{eq:HpRepresentation} & $0.5$ & $[10^{-4}, 10^{-2}]$ & $[10^{-6}, 10^{-3}]$ \\ 
\hline 
PSR w/o~\eqref{eq:HpRepresentation} & $1.0$ & $[10^{-4}, 10^{-2}]$ & $[10^{-6}, 10^{-3}]$ \\
\hline
PSR w/o~\eqref{eq:HpRepresentation} & $2.0$ & $[10^{-4}, 10^{-2}]$ & $[10^{-6}, 10^{-4}]$ \\ 
\hline
\end{tabular}
\end{center}
\end{table}

\begin{table}[h]
\caption{CNN Test Loss and Accuracy on CIFAR-10}
\label{table:CNNPerformance}
\begin{center}
\begin{tabular}{|c|c|c|c|c|}
\hline
Algorithm & RS 2x & SH & HB & PGSR-HB \\
\hline
Loss (I) & $0.7118$ & $0.7001$ & $0.7150$ & $\mathbf{0.6455}$ \\
Acc (I) & $81.17\%$ & $79.69\%$ & $78.74\%$ & $\mathbf{82.79\%}$ \\
\hline
Loss (II) & $0.6988$ & $0.7179$ & $0.6921$ & $\mathbf{0.6764}$ \\
Acc (II) & $79.51\%$ & $79.30\%$ & $81.67\%$ & $\mathbf{83.00\%}$ \\
\hline
Loss (III) &$0.6850$& $0.6747$& $0.6960$& $\mathbf{0.6467}$\\
Acc (III) &$79.02\%$& $79.80\%$& $\mathbf{81.47\%}$& $80.39\%$\\
\hline
Loss (IV) &$0.7293$&$\mathbf{0.6499}$&$0.7215$&$0.6619$ \\
Acc (IV) &$77.70\%$&$80.68\%$&$80.81\%$&$\mathbf{81.64\%}$\\
\hline
\end{tabular}
\end{center}
\end{table}


Next, we optimize the five categories of hyperparameters including the learning rate, three convolution layers' and a fully connected dense layer's Tikhonov regularization constants using the same architecture and dataset used in the previous section. We trained the network using the stochastic gradient descent without a momentum and included the learning rate decay by a factor $0.1$ every 100 epochs of training. We compare SH, Hyperband, Random Search with doubled budgets and PGSR-HB based on test loss and accuracy. We set the resource $R = 243$ and the discard ratio input $\eta=3$ and allocated the equivalent total budget between the algorithm based on the training epochs except for Random Search 2x. Setting the total budget of four cycles of Hyperband and PGSR-HB as the baseline, Random Search 2x evaluates 288 randomly sampled hyperparameter configurations with the resource $R$ and SH cycles 24 times as one Hyperband contains six subroutine SH. Since the randomness involves in these hyperparameter optimization algorithms, we compare four different trials of each algorithms as shown in Table~\ref{table:CNNPerformance}. The experiment result verifies the effectiveness of reducing the hyperparameter space through PGSR as the new algorithm returns better performance for most of the trials. Moreover, PGSR-HB found the optimal hyperparameters returning $83\%$ test accuracy which outperforms the other algorithms from all trials.

\section{Conclusion}
\label{sec:con}

We proposed a new HPO algorithm which learns the most influential hyperparameters by carefully tracking loss function (measurement) history in a Hyperband framework. Our new algorithm is based on a key modification of polynomial sparse recovery (PSR) that induces further improvement via a group-sparsity constraint. Future directions include performing a multi-stage Group Lasso to reduce hyperparameter space further as we obtain new observations. While the goal of the HPO problem is to approximate the global minimizer of the loss over hyperparameter space, HPO methods themselves require tuning, so a fully automatic ML training method is still of great interest.
{{
\bibliographystyle{IEEEbib}
\bibliography{chomdbiblio}
}}
\end{document}